\documentclass[pmlr]{jmlr}
\usepackage{subcaption}
\usepackage{algorithm}
\usepackage{algpseudocode}

\RequirePackage{graphicx}
 \usepackage{booktabs}
\usepackage{longtable}
\makeatletter
\def\set@curr@file#1{\def\@curr@file{#1}} 
\makeatother
\usepackage[load-configurations=version-1]{siunitx} 

\theorembodyfont{\upshape}
\theoremheaderfont{\scshape}
\theorempostheader{:}
\theoremsep{\newline}

\jmlrvolume{tbd}
\jmlryear{2025}
\jmlrworkshop{International Conference on Computational Optimization}

\title[SCOPE]{SCOPE: Smooth Convex Optimization for Planned Evolution of Deformable Linear Objects}

\author{%
  \Name{Ali Jnadi} \Email{a.jnadi@innopolis.university}\\
  \addr Phystech School of Applied Mathematics and Computer Science, MIPT, Russia\\
  Research Center for Artificial Intelligence, Innopolis, Russia\\
  Q Deep, Innopolis , Russia
  \AND
  \Name{Hadi Salloum} \Email{h.salloum@innopolis.ru}\\
  \addr Phystech School of Applied Mathematics and Computer Science, MIPT, Russia\\
  Research Center for Artificial Intelligence, Innopolis University, Russia\\
  Q Deep, Innopolis , Russia
\AND
  \Name{Yaroslav Kholodov} \Email{y.kholodov@innopolis.ru}\\
  \addr Phystech School of Applied Mathematics and Computer Science, MIPT, Russia\\
  Laboratory of Quantum Computing, Innopolis University, Russia
  \AND
  \Name{Alexander Gasnikov} \Email{gasnikov@yandex.ru}\\
  \addr Phystech School of Applied Mathematics and Computer Science, MIPT, Russia\\
  Research Center for Artificial Intelligence, Innopolis University, Russia
  \AND
  \Name{Karam Almaghout} \Email{k.almaghout@innopolis.university}\\
  \addr Center for Technologies in Robotics and Mechatronics Components, Innopolis University, Russia
}

\begin{document}

\maketitle
  
\begin{abstract}
 We present SCOPE, a fast and efficient framework for modeling and manipulating deformable linear objects (DLOs). Unlike conventional energy-based approaches, SCOPE leverages convex approximations to significantly reduce computational cost while maintaining smooth and physically plausible deformations. This trade-off between speed and accuracy makes the method particularly suitable for applications requiring real-time or near-real-time response. The effectiveness of the proposed framework is demonstrated through comprehensive simulation experiments, highlighting its ability to generate smooth shape trajectories under geometric and length constraints.
\end{abstract}

\section{Introduction}
DLOs, such as cables, ropes, and surgical sutures, are ubiquitous in industrial, medical, and domestic applications \cite{trommnau2019overview,heisler2017automatization,xu2018real,costanzo2023enhanced,sanchez2018robotic}. Their manipulation is challenging due to their high degrees of freedom, nonlinear deformation, and sensitivity to contact, which makes both modeling and control computationally demanding \cite{salloum2024enhancing, zhu2022challenges,yin2021modeling,arriola2020modeling}. This difficulty has motivated a wide range of approaches, including finite element methods \cite{duenser2018interactive,koessler2021efficient}, mass-spring models \cite{lv2017physically,patete2013multi}, elastic rod formulations \cite{linn2017discrete,valentini2011modeling}, and more recent learning-based strategies \cite{yan2020self,yang2022learning,wang2022offline}. While these models can achieve accurate deformations, their computational cost often limits applicability in real-time robotic tasks such as cable routing, assembly, and surgical suturing \cite{fresnillo2022approach,laezza2021learning,zakaria2022robotic}.

A recent line of research introduced energy-based methods for DLO modeling and manipulation, which discretize the object into rigid links connected by elastic elements and compute equilibrium shapes by minimizing internal energy \cite{almaghout2023deformable,almaghout2022planar}. These methods capture physical behavior and yield stable solutions, but solving the associated nonlinear optimization is computationally expensive, requiring seconds to minutes depending on the task. This makes them unsuitable for real-time applications such as closed-loop shape control \cite{lagneau2020automatic,ruan2018accounting} or simulation-intensive reinforcement learning \cite{laezza2021shape,zakaria2022robotic}. These limitations make computation time a major bottleneck in practical applications, particularly in scenarios that demand fast responses or repeated simulations, such as robotic inspection tasks \cite{salloum2024quantum}.

To address this limitation, we propose SCOPE, a novel approximate framework that replaces nonlinear energy minimization with a convex optimization problem. Unlike prior formulations that explicitly model stretching and bending energies, SCOPE enforces essential physical constraints, such as in-extensibility and smoothness, through simple geometric terms and quadratic costs. This significantly reduces computation time while preserving sufficient fidelity for robotics applications. Our evaluation on 2D shape transformation tasks shows that SCOPE achieves comparable accuracy to energy-based methods but with order-of-magnitude faster solve times, making it more practical for real-time control and learning-based pipelines.

The rest of this paper is structured as follows: \sectionref{2} introduces the preliminaries and problem statement. \sectionref{3} presents the proposed method. \sectionref{4} describes simulation results. \sectionref{5} concludes the paper and discusses future work.

\section{Preliminaries}
\label{2}

\begin{figure}[htbp]
    \centering

    \subfigure[DLO mass-spring model]{
        \includegraphics[width=0.4\textwidth]{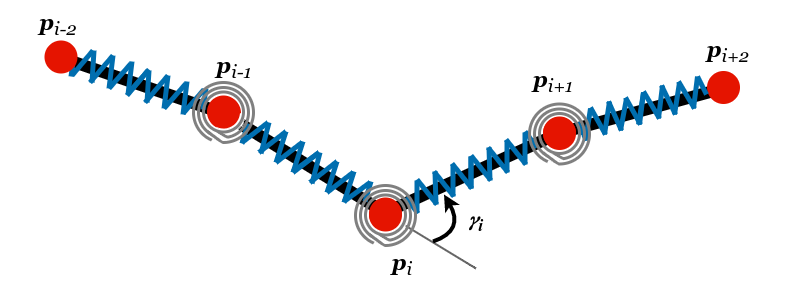}
        \label{fig:Model}
    }
    \subfigure[Schematic of the \textcolor{orange}{robots end-effectors}, 
               the \textcolor{gray}{cable}, and the 
               \textcolor{green}{desired shape}.]{
        \includegraphics[width=0.4\textwidth]{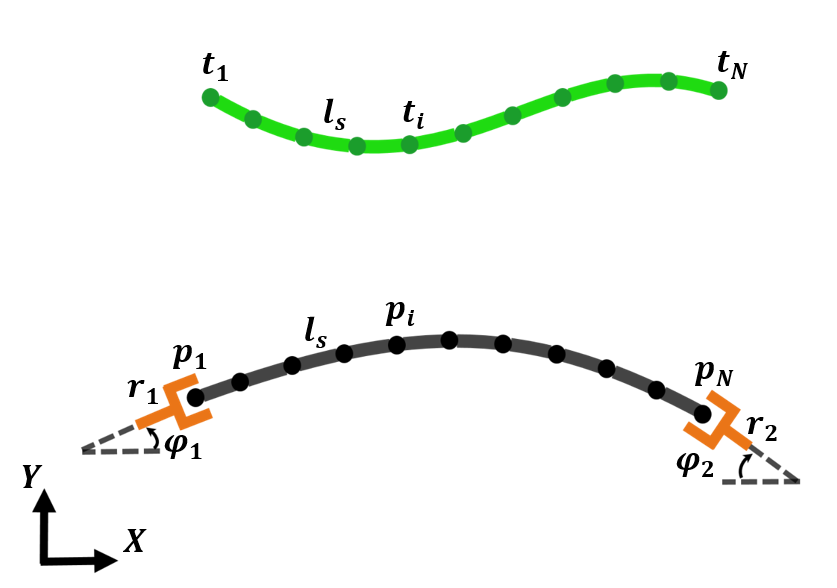}
        \label{fig:Task}
    }

    \caption{Model and Application.}
    \label{fig:mainfig}
\end{figure}

We consider a planar deformable linear object of fixed length, modeled as a sequence of $N$ point (nodes) $p_1, p_2, \dots, p_N$ connected by $N - 1$ segments, as shown in  \figureref{fig:Model}. The endpoints $p_1$ and $p_N$ correspond to the ends of the cable and are assumed to be controllable (grasped by robotic grippers), while the intermediate points ${p_i}\ \text{for}\ i=2, \cdots,\ N-1$ passively move, as shown in \figureref{fig:Task}. 

In an energy-based framework, the DLO deformability is captured by assigning elastic energies to stretching and bending of the segments. Each segment between $p_{i}$ and $p_{i+1}$ has an associated stretching (extensional) energy $E_s^i$, which penalizes changes in length from the rest length $l_s$ (the nominal segment length when the rope is slack). Likewise, each adjacent pair of segments connected by an internal node $p_i$ for $2\le i\le N-1$ has a bending (torsional) energy $E_b^i$ that penalizes deviations of the angle between those segments from the initial angle. For example, a simple model can define:
\begin{equation}
    E_s^i = \frac{1}{2}k_s(|p_{i+1}-p_i| - l_s)^2
\end{equation}
and:
\begin{equation}
E_b^i = \frac{1}{2}k_b (\gamma_i - \gamma_1)^2
\end{equation}
where $k_s$ and $k_b$ are stretching and bending stiffness coefficients, and $\theta_i$ is the turning angle at node $p_i$. Thus, the total internal energy of the DLO, $E_{\text{internal}}$, represents the potential energy stored in deformations of the object, and given as follows:
\begin{equation}
    E_{\text{internal}} = \sum_{i=1}^{N-1} E_s^i + \sum_{i=2}^{N-1} E_b^i
\end{equation}

Under the energy-based approach, the static shape of a DLO for given end-point positions is determined by minimizing the internal energy $E_{\text{internal}}$ subject to the geometric constraint of fixed end positions and (optionally) inextensible length. In essence, the DLO will assume the configuration that balances stretching and bending forces, this is computed by solving a constrained optimization, or equivalently, finding the equilibrium of forces derived from the energy. Analytical solutions are generally intractable for arbitrary boundary conditions, so numerical methods like nonlinear programming or gradient descent on the energy landscape are used. The energy function is typically nonconvex due to the bending terms (which depend on angles or curvature), requiring careful initialization to converge to a physically correct configuration.

To drive a DLO from an initial shape to a desired target shape, the prior work, ~\cite{almaghout2023deformable, almaghout2024manipulation}, formulated a trajectory optimization problem that incorporates the internal energy into the objective. One convenient strategy is to introduce a cost term for shape error alongside the energy. For instance, given a desired curve (target shape), one can define an objective:
\begin{equation}
    J = E_{\text{internal}}(p_0,\dots,p_N) + \lambda E_{\text{shape}}(p_0,\dots,p_N)
\end{equation}
where $E_{\text{shape}}$ measures the discrepancy between the current DLO configuration and the target configuration (e.g. sum of squared distances from each $p_i$ to its corresponding target point on the desired curve). The weight $\lambda$ tunes the trade-off between strictly following the target shape and keeping the DLO in a low-energy (physically plausible) state. Additional constraints or penalty terms are included to maintain the DLO’s characteristics; for example, a hard or soft constraint to preserve total length (preventing excessive stretch) and perhaps a regularization to avoid erratic shapes. Solving this optimization yields the final positions of all points $p_i$ that best achieve the target shape while respecting physical realism. In the energy-based method, this optimization problem is generally nonlinear and can be high-dimensional, so solving it often involves iterative numerical solvers that might be slow. Indeed, while this approach achieves accurate results and stable deformations, its computational cost grows quickly with the number of segments and the complexity of the shape change. As we show later, the baseline method can take tens of seconds up to solve for a single shape transition, which is impractical for time-critical applications. Moreover, implementing the energy-based optimizer requires handling complex derivatives of the energy and carefully tuning parameters like stiffness coefficients for different DLO materials.

These challenges motivate a faster, more streamlined approach. In what follows, we present our SCOPE method, which forgoes direct computation of energy terms in favor of a convex approximation. We will reuse the above notation and physical intuition but simplify the formulation to gain efficiency.

\section{Methodology}
\label{3}


The proposed method computes a feasible deformation trajectory from the initial shape to the target shape by solving a convex optimization problem. We discretize the deformation process into $T$ time steps (including the start at $t=1$ and the goal at $t=T$) and simultaneously solve for the intermediate configurations of the DLO at each time step. The decision variables can be represented as $p_{\text{var}}(i, t)$, which gives the 2D coordinates of the $i$-th point of the DLO at intermediate stage $t$. In other words, $p_{\text{var}} \in \mathbb{R}^{N\times 2 \times T}$ represents the entire trajectory of the DLO (all node positions over all time steps). The optimization is initialized with known values for the endpoints at the first and last time steps: we constrain $p_{\text{var}}(0,t) = p_0^{(t)}$ and $p_{\text{var}}(N,t) = p_N^{(t)}$ such that $p_{\text{var}}(:,1)$ equals the coordinates of the initial shape and $p_{\text{var}}(:,T)$ equals the coordinates of the desired final shape. (Here $p_{\text{var}}(:,t)$ denotes the set of all point coordinates at time $t$.) The intermediate configurations $p_{\text{var}}(:,2), ..., p_{\text{var}}(:,T-1)$ are initially unconstrained and will be determined by the optimizer.

Inextensibility Constraints: To prevent unphysical stretching of the DLO, we enforce a hard constraint on the length of each segment at every time step. Specifically, for each segment between consecutive points $p_{\text{var}}(i,t)$ and $p_{\text{var}}(i+1,t)$, we impose

\begin{equation}
    p_{var}(i+1,:,j) - p_{var}(i,:,j)\|_2 \leq l_0, \quad \forall i=0, \cdots, N-1, \forall t = 1, \cdots, T
\end{equation}

where $l_0$ is the rest (or maximum allowable) length of each segment. This ensures the cable does not stretch beyond its original length. 
Instead of explicitly formulating a bending energy, we encourage smooth deformations by penalizing differences between successive time-step configurations. Intuitively, if the DLO shape changes gradually from one time step to the next, it implies that no large bending or sudden motion occurs at once. We introduce a smoothness cost defined as the sum of squared displacements of each point between consecutive time steps. Let $p_{\text{var}}^x(i,t)$ and $p_{\text{var}}^y(i,t)$ denote the $x$ and $y$ coordinates of point $i$ at time $t$. We define:

\begin{equation}
    \begin{aligned}
        x_{obj} = \sum_{j=1}^{N-1} \| p_{var}(:,1,j) - p_{var}(:,1,j+1) \|^2
\\
y_{obj} = \sum_{j=1}^{N-1} \| p_{var}(:,2,j) - p_{var}(:,2,j+1) \|^2
    \end{aligned}
\end{equation}
where $x_{obj}$ and $y_{obj}$ measure how much each point $x$ and $y$ positions move from one step to the next, summed over all points and all time intervals. The total smoothness objective can be taken as:
\begin{equation}
    S_{obj} = x_{obj} + y_{obj} = \sum_{t=1}^{T-1}\sum_{i=0}^{N} | p_{var}(i,t+1) - p_{var}(i,t) |_2^2
\end{equation}
which directly penalizes rapid changes. This encourages the optimizer to find a sequence of shapes that changes gradually, effectively approximating the effect of bending stiffness without explicitly computing curvature, large bends would require big position changes in one step and thus incur a high cost, so the optimizer prefers many small bends spread across steps (a smoother curve).

We found that adding a guiding trajectory for the intermediate shapes improves convergence and solution quality. Before optimization, we construct a set of intermediate guide points (or even full intermediate shapes) denoted as $inter\_pts(:, :, t)$ for $t=2,\dots,T-1$. These could be obtained by a simple interpolation between the initial and final shapes or by any other heuristic that provides a rough path from start to goal. For example, one could linearly interpolate each point’s position from the start to end configuration over the $T$ steps, resulting in a straight-line motion for each point; this would yield a reasonable intermediate shape for each $t$ that smoothly morphs the DLO from start to finish. We then introduce a midpoint penalty term that keeps the optimized intermediate shape close to these guide points:

\begin{equation}
    M_{obj} = \sum_{t=1}^{T-1}\sum_{j=2}^{N-1} \| p_{var}(:,:,j) - inter\_pts(:,:,j) \|_2
\end{equation}

This term is minimized when the solution’s intermediate configurations coincide with the chosen guide shapes. By tuning the weight of this term, we can allow the optimizer freedom to deviate from the initial guess if needed, but still gently pull the solution toward a sensible path. The midpoint term serves a role analogous to the energy-based model’s tendency to favor “natural” configurations – it helps avoid highly contorted or inefficient deformation paths by providing a reasonable prior trajectory.

The final optimization problem combines the smoothness objective and the midpoint guidance, with weighting factors to balance their influence. We define the total cost function as:

\begin{equation}
    J(p_{var}) = w_1S_{obj} + w_2M_{obj}
\end{equation}
where $w_1$ and $w_2$ are nonnegative weights chosen according to the desired trade-off. A typical choice might be to weight smoothness relatively high to ensure physical plausibility, and use a smaller weight on the midpoint term just to provide gentle guidance (unless the user wants to enforce following a very specific intermediate path). Importantly, we do not explicitly include a term for final shape error in $J$. The final shape matching is instead guaranteed by the hard constraint that $p_{var}(:,T)$ equals the desired shape. In other words, our method explicitly constrains the start and end configurations to be exactly the ones specified, rather than treating the end-shape error as a soft cost. This is a deliberate design choice: since we allow a small violation of physical optimality, we ensure that the end goal is met exactly (or within solver tolerance) rather than approximately. In contrast, the energy-based method might end up with a tiny residual shape error if it balances against energy, but here we force zero error at $t=T$ by constraint. The cost function $J$ therefore only governs how the DLO moves through the intermediate steps. Once $J$ is defined, the SCOPE is formulated as:

\begin{equation}
    \begin{aligned}
        & \underset{p_{var}}{\text{maximize}} & & J(p_{\text{var}}) \\
        & \text{subject to:}
        & &\begin{cases}
            p_{{var}}(:,1) = \text{InitialShape} \\
            p_{{var}}(:,T) = \text{TargetShape} \\
            \|p_{{var}}(i+1,t) - p_{{var}}(i,t)\|_2^2 \le l_0, \forall (i, t) = ((0,\dots,N-1),\ (1,\dots,T))
    \end{cases}
    \end{aligned}
\end{equation}

This is a convex optimization problem: $J$ is a quadratic function in the decision variables, and the constraints are linear equalities or convex inequalities. We solve it using a standard convex solver. In our implementation, we used the CVX framework to model and solve the problem efficiently. The solver returns the optimal positions $p_{{var}}(i,t)$ for all points and time steps. From these, we obtain a trajectory of DLO shapes that smoothly transition from the start shape to the goal shape, as illustrated in Algorithm~\ref{alg:shape_opt}.

Comparison to energy-based formulation, it is illuminating to contrast how our SCOPE method differs from the energy-based baseline in formulation and complexity. The energy-based approach requires calculating forces/energies related to curvature (bending) and extension, which typically involves nonlinear functions (e.g., trigonometric functions for angles or polynomial terms for stretch). Optimizing those directly leads to a nonlinear program that may have multiple local minima and demands iterative solvers. In our approach, we replaced bending energy with a surrogate quadratic smoothness cost and replaced stretching energy with a simple length constraint. Both of these substitutions greatly simplify the mathematics: smoothness is now a convex quadratic term, and inextensibility is a convex constraint, meaning the entire problem is convex. As a result, global optimality of the solution is guaranteed (the solver finds the true minimum of $J$) and the computation can exploit highly optimized algorithms for convex problems. This yields a faster solution without getting trapped in poor local minima. Moreover, implementation is simpler – one does not need to derive gradients of complicated energy expressions or simulate physics; instead, one only needs to set up difference calculations and norm constraints, which is straightforward in tools like CVX. The trade-off is that $J$ is no longer a physical energy – it does not rigorously model material stiffness. Thus, the obtained intermediate shapes might not exactly follow physical laws (e.g. the DLO might sag less in the absence of gravity or might cut corners slightly because we did not include an exact curvature integral). However, as we will show, the resulting paths are still quite plausible and the final shape can be reached with only a minor error. The design of $J$ was chosen to retain the qualitative behavior of a stiff DLO (preferring smooth transitions and not stretching), which in practice suffices for many applications.

\begin{algorithm}[h]
\caption{SCOPE: Smooth Convex Optimization for Planned Evolution of Deformable Linear Objects}
\label{alg:shape_opt}
\begin{algorithmic}[1]
\State \textbf{Input:} Start shape $P_s$, target shape $P_e$, number of points $N$, time steps $T$, segment length $l_s$, intermediate guide points $P_{intmd}$, weights $w_1, w_2$
\State \textbf{Initialize} decision variable $p_{var} \in \mathbb{R}^{N \times 2 \times T}$
\State \textbf{Set boundary conditions:} 
\[
p_{var}(:,:,1) = P_s, \quad p_{var}(:,:,T) = P_e
\]
\State \textbf{Enforce inextensibility constraints:} 
\[
\|p_{var}(i+1,:,t) - p_{var}(i,:,t)\|_2 \le l_s, \quad \forall i=1\dots N-1, \ \forall t=1\dots T
\]
\State \textbf{Define smoothness objective:} 
\[
S_{obj} = \sum_{t=1}^{T-1}\sum_{i=1}^{N} \| p_{var}(i,t+1) - p_{var}(i,t) \|_2^2
\]
\State \textbf{Define midpoint guidance penalty:} 
\[
M_{obj} = \sum_{t=2}^{T-1}\sum_{i=2}^{N-1} \| p_{var}(i,t) - P_{intmd}(i,t) \|_2^2
\]
\State \textbf{Form total cost function:} 
\[
J = w_1 S_{obj} + w_2 M_{obj}
\]
\State \textbf{Solve the convex optimization problem to obtain $p_{var}$}
\State \textbf{Output:} Optimized DLO trajectory $p_{var}(:,:,1:T)$
\end{algorithmic}
\end{algorithm}

Finally, it is worth noting that our algorithm naturally produces a trajectory of shapes (the sequence $p_{{var}}(:,t)$ for $t=1$ to $T$). This is in contrast to some energy-based formulations which might only compute the final static shape or require a separate procedure (e.g. quasi-static stepping) to generate a path. By formulating time explicitly, we ensure the solution is a continuous deformation process. This is advantageous for robotic execution: the sequence of intermediate shapes can be used to command the robot in small increments, ensuring a smooth motion that respects cable constraints throughout the manipulation (rather than suddenly imposing the final shape, which could cause large impulsive forces). In summary, the SCOPE provides a fast, convex, and easy-to-implement alternative to energy-based DLO modeling, suitable for scenarios where a slight approximation is acceptable for the gain in speed.

\section{Simulation Results}
\label{4}
\begin{figure}[!ht]
    \centering
    \includegraphics[width=\linewidth]{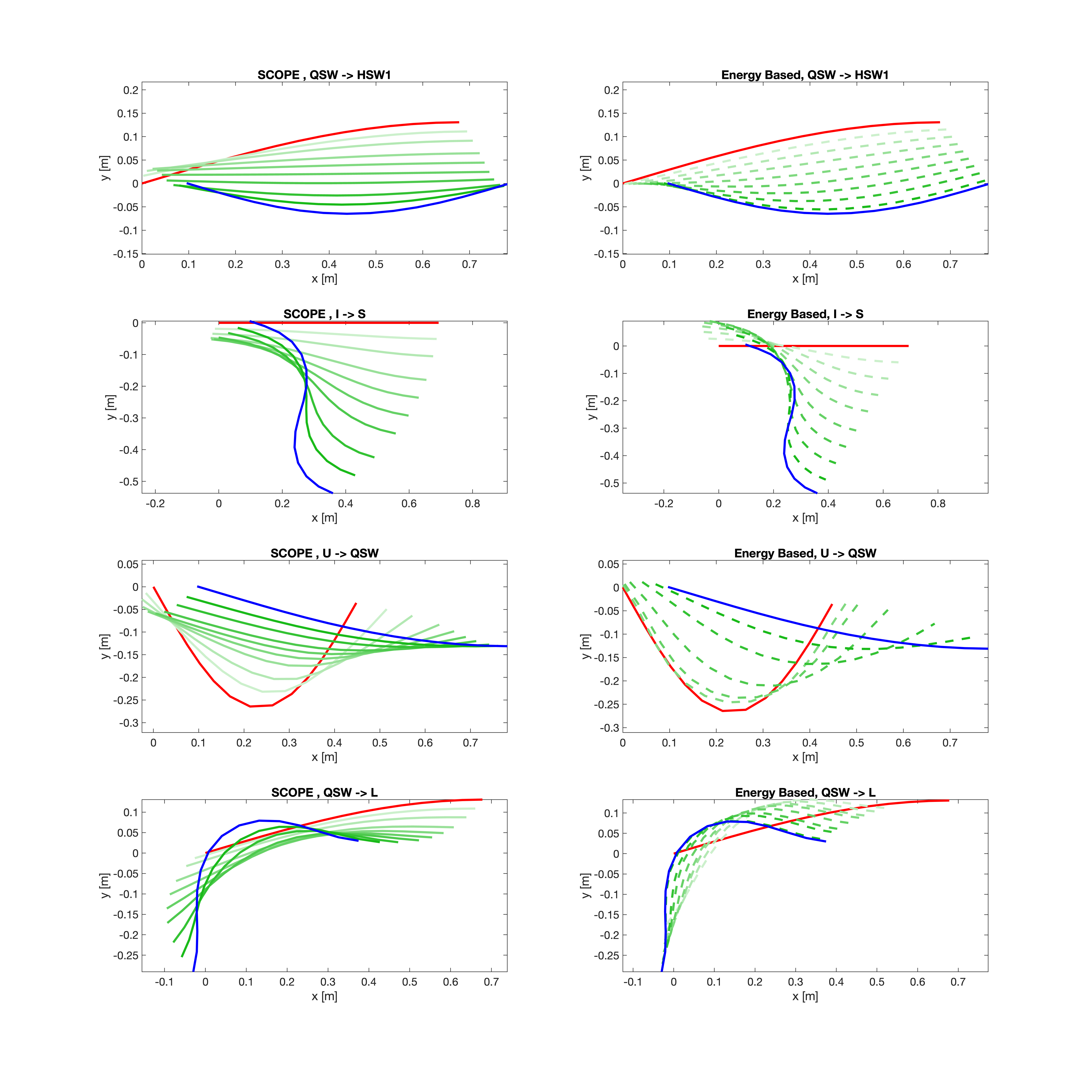}
    \caption{Results of the optimization experiments. 
    Each row corresponds to a different desired shape conversion, as indicated in the subplot titles. 
    The \textcolor{red}{red solid curve} represents the initial cable configuration, while the \textcolor{blue}{blue solid curve} indicates the final target shape. 
    The \textcolor{green}{solid green trajectories} correspond to the intermediate deformations obtained with shape optimization, while the \textcolor{green}{dashed green trajectories} correspond to the intermediate deformations obtained with energy-based optimization. 
    This visualization highlights the differences in convergence behavior between the two approaches.}
    \label{fig:placeholder}
\end{figure}

We evaluated the proposed SCOPE against the baseline energy-based method on a variety of planar shape transformation tasks. Each task involves an initial DLO configuration and a desired final configuration; these configurations include shapes with varying curvature and complexity \footnote{Numerical analyses were performed in MATLAB\textregistered~{}R2024a Update 4. The energy-based model was optimized using the \texttt{fmincon} solver from the Optimization Toolbox. SCOPE was implemented using the CVX modeling framework~\cite{cvx, gb08} and solved with the \texttt{SDPT3} solver (version 4.0). All computations were executed on a workstation running Microsoft Windows\textregistered~{}10 Pro, equipped with an Intel\textregistered{} Core\texttrademark{} i7-9700K CPU (8 cores, 3.60~GHz), 32.0~GB of RAM, and an NVIDIA\textregistered{} GeForce RTX 2060 GPU (6.0~GB VRAM, CUDA compute capability 7.5).}. For brevity we label the shapes with shorthand names: "QSW" denotes a shape resembling a quarter-sine wave, "HSW" a half-sine wave, "U" a U-shaped curve, "S" an S-shaped curve, "I" an I-shaped curve, and "L" an L-shape. Figure~\ref{fig:placeholder} provides a visual illustration of four representative experiments. In each subfigure, the red curve shows the DLO initial shape and the blue curve shows the target shape. The green curves in between are the intermediate configurations produced by our SCOPE method, while the dashed green one is the configurations produced by SCOPE. As seen in Figure~\ref{fig:placeholder}, the method produces smooth and continuous deformations from start to finish, there are no abrupt jumps or unphysical kinks. This validates that the smoothness objective successfully regularizes the trajectory, and keep the deformation on a reasonable path.

\begin{table}[htbp]
\centering
\caption{Performance of SCOPE vs.\ Energy-Based Optimization (time: s, error: m)}

\begin{tabular}{|c|c|c|c|c|c|}
\hline
Start Shape & End Shape  & \multicolumn{2}{c|}{SCOPE} & \multicolumn{2}{c|}{Energy Based} \\
\cline{3-6}
            &            & Solve Time & Max Error & Solve Time & Max Error \\
\hline
QSW & HSW & \textbf{2.4} & \textbf{0.7} & 22.73 & \textbf{0.7} \\
\hline
I   & S    & \textbf{1.52} & 5.2  & 59.13 & \textbf{0.7} \\
\hline
U   & QSW  & \textbf{1.4} & 1.6  & 31.91 & \textbf{0.8} \\
\hline
QSW & L    & \textbf{3.01} & 2.8  & 187.52 & \textbf{0.7} \\
\hline
\end{tabular}

\label{tab:shape_solutions}
\end{table}

Table~\ref{tab:shape_solutions} (Time in \textbf{seconds} and error in \textbf{meter}) summarizes the performance of the SCOPE versus the Energy-Based approach on four different shape transitions. The metrics reported are the solve time (optimization runtime in seconds) and the maximum length error within all configurations (in centimeters). The maximum shape error is defined as the largest difference between the length of any configuration and the length of DLO; essentially it captures the worst-case as compress in the DLO which could lead to bending or unintended motion on z direction. For the energy-based method, this error is expected to be very small since that method directly optimizes shape accuracy (at the cost of more computation). Indeed, in our experiments the energy-based approach achieved final errors on the order of a few millimeters (effectively zero at the scale of the shapes considered). Our accelerated method, by contrast, sometimes yields a small residual error because it emphasizes speed and smoothness over absolute accuracy.

As the table shows, our method consistently outperforms the baseline in speed, often by a large margin. For moderate shape changes like QSW-to-HSW, SCOPE found a solution in about 2.74 seconds compared to 22.73 seconds for the energy method (approximately 8 times faster). In more challenging cases with substantial reconfiguration, QSW-to-L-shape, which requires flipping a half-wave into a right-angle shape, SCOPE took 187.52 seconds to converge, whereas SCOPE needed only 4.03 seconds. This is a speed-up by a factor of 47 times, dramatically highlighting the benefit of our approach for complex tasks. Even in the I to S case, which involves significant bending (an initially straight DLO into an S curve), our method solved in roughly 3.0 seconds versus 59.13 seconds for the baseline. We note that the solve times for SCOPE include the overhead of the convex solver but are dominated by the problem size (number of variables); in all tested scenarios, the runtime remained on the order of a few seconds, showing good scalability and potential for near-real-time use if further optimized.

In terms of accuracy, the SCOPE achieved small to moderate shape errors depending on the scenario. For two of the shape transitions (QSW-to-HSW and U-to-QSW), the final maximum error was under 2.0cm, which is quite negligible (the DLO length in those cases is on the order of 70.0cm, so $<3\%$ relative error). In these cases, the approximate method’s result was almost as good as the exact energy-minimizing result, but obtained much faster. The QSW-to-HSW case, for instance, had only 0.7 cm error with our method. The energy-based solution in those cases had an error $<10^{-1}cm$. The largest observed error for our method was 5.2cm in the I-to-S transformation. This is a noticeable deviation, reflecting that the S-shaped target has high curvature sections that our faster method lead to some compressing to achieve. However, even 5.2cm error may be acceptable in many practical contexts (depending on the application’s tolerance), and it comes with the trade-off of a 20 times speed improvement. Moreover, it is possible to reduce this error by increasing the number of intermediate steps $T$ or adjusting weights, in essence, allowing the DLO more opportunity to less compressing at the expense of additional computation. In all cases, the energy-based method error was essentially less than $<10^{-1}cm$, since it explicitly minimizes that. Therefore, although the energy method is slightly more accurate, but our method is vastly faster.

Beyond quantitative metrics, the two approaches exhibit salient qualitative differences. SCOPE was straightforward to implement: it avoids differentiating energy functionals and simulating dynamics, relying instead on a high-level convex modeling tool. This yields a compact implementation with reliable convergence. The energy-based method required careful parameter tuning and occasionally exhibited convergence issues on complex targets; the QSW-to-L instance, for example, produced extended runtimes consistent with many iterations or a challenging energy landscape. These properties affect practical use. A fast planner enables closed-loop control in which a robot continuously reshapes a cable to track a moving target. The same advantages carry over to reinforcement learning and high-level planning, where large batches of DLO simulations are needed.

SCOPE sacrifices a small amount of accuracy for speed, a trade-off that is acceptable in many settings. When exactness is critical, a hybrid procedure can be employed: use SCOPE to generate an initial guess and warm-start an energy-based refinement. In our experiments, SCOPE was sufficiently accurate for the majority of tasks.

The trajectories generated by the proposed method are smooth, which facilitates execution. The intermediate states in Figure~\ref{fig:placeholder} exhibit gradual shape evolution that yields low-variation velocity profiles for the manipulators. By contrast, the energy-based approach, when used for trajectory generation, does not promote smoothness unless time parameterization and explicit regularization are introduced. Our objective explicitly includes smoothness terms, producing gradual paths by construction. This confers a practical advantage for hardware control by mitigating impulsive forces and abrupt configuration changes.

\section{Conclusion}
\label{5}
This study compares two optimization approaches for deformable linear objects: the SCOPE method and a traditional energy-based formulation. Both approaches successfully achieved the target configurations across four test cases, but their performance characteristics differ markedly. The energy-based optimization consistently delivers higher accuracy, albeit with slower convergence, whereas the SCOPE method is significantly faster but may struggle to achieve the same precision, particularly for more complex target shapes. These complementary characteristics highlight that method selection should be guided by application requirements. The energy-based formulation is preferable in scenarios where high precision is critical, while the SCOPE method offers advantages in time-sensitive applications or as a warm-start strategy for more accurate solvers. Future work will explore extensions to more complex geometries, three-dimensional deformable objects, and real-world experiments, as well as hybrid strategies that balance computational speed and solution accuracy.

\acks{The study was supported by the Ministry of Economic Development of the Russian Federation (agreement No. 139-10-2025-034 dd. 19.06.2025, IGK 000000C313925P4D0002).}

\bibliography{sample}






\end{document}